\documentclass[letterpaper, 10 pt, conference]{ieeeconf}  %

\IEEEoverridecommandlockouts                              %

\usepackage{cite}
\usepackage{hyperref}
\usepackage{url}

\usepackage[utf8]{inputenc} %
\usepackage[T1]{fontenc}    %
\usepackage{hyperref}       %
\usepackage{url}            %
\usepackage{booktabs}       %
\usepackage{amsfonts}       %
\usepackage{nicefrac}       %
\usepackage{microtype}      %
\usepackage[table,xcdraw]{xcolor}

\usepackage{csquotes}
\MakeOuterQuote{"}
\usepackage{graphicx}
\let\labelindent\relax
\usepackage{enumitem}
\usepackage{algorithm}
\usepackage[noend]{algpseudocode}
\usepackage{amsmath}
\usepackage{amssymb}
\usepackage{mathtools}
\usepackage{wrapfig}
\usepackage[normalem]{ulem}

\usepackage[font=small]{subcaption}
\usepackage[font=small,labelfont=bf]{caption}

\usepackage{listings}
\usepackage{xcolor}

\definecolor{codegreen}{rgb}{0,0.6,0}
\definecolor{codegray}{rgb}{0.5,0.5,0.5}
\definecolor{codepurple}{rgb}{0.58,0,0.82}
\definecolor{backcolour}{rgb}{0.95,0.95,0.92}

\newcommand{\data}{\mathcal{D}}

\newcommand{\policy}{\pi}

\title{\LARGE \bf
AdaDemo: Data-Efficient Demonstration Expansion \\ for Generalist Robotic Agent
}

\author{
Tongzhou Mu$^{2, *}$, Yijie Guo$^{1}$, Jie Xu$^{1}$, Ankit Goyal$^{1}$, Hao Su$^{2}$, Dieter Fox$^{1,3}$, Animesh Garg$^{1,4}$
\thanks{*Work is done at NVIDIA.}%
\thanks{$^{1}$NVIDIA, $^{2}$UC San Diego, $^{3}$University of Washington, $^{4}$Georgia Tech.}%
}

\begin{document}

\maketitle
\thispagestyle{empty}
\pagestyle{empty}

\setlist*[itemize]{labelindent=\parindent,itemindent=0pt,leftmargin=10pt}

\begin{abstract}

Encouraged by the remarkable achievements of language and vision foundation models, developing generalist robotic agents through imitation learning, using large demonstration datasets, has become a prominent area of interest in robot learning. The efficacy of imitation learning is heavily reliant on the quantity and quality of the demonstration datasets. 
In this study, we aim to scale up demonstrations in a data-efficient way to facilitate the learning of generalist robotic agents. We introduce AdaDemo (Adaptive Online Demonstration Expansion), a general framework designed to improve multi-task policy learning by actively and continually expanding the demonstration dataset. AdaDemo strategically collects new demonstrations to address the identified weakness in the existing policy, ensuring data efficiency is maximized. 
Through a comprehensive evaluation on a total of 22 tasks across two robotic manipulation benchmarks (RLBench and Adroit), we demonstrate AdaDemo’s capability to progressively improve policy performance by guiding the generation of high-quality demonstration datasets in a data-efficient manner.

\end{abstract}
    
\section{Introduction}
\label{sec:intro}

The recent unprecedented success of language and vision foundation models \cite{brown2020language, kirillov2023segment} have highlighted the importance of scaling up datasets as a key strategy for solving challenging tasks. 
This insight has similarly influenced the field of robotics, where the development of robotic foundation models through imitation learning, utilizing large datasets of demonstrations, has emerged as a prominent area of interest \cite{bousmalis2023robocat, brohan2022rt, brohan2023rt, ahn2022can}. 
However, the success of these robotic foundation models is significantly dependent on the quantity, quality, and diversity of the demonstration data. 
Meanwhile, recent efforts of collecting large-scale robot demonstration datasets, such as RT-1 \cite{brohan2022rt} and Open X-Embodiment \cite{open_x_embodiment_rt_x_2023}, require extensive time and resources, involving years of data collection by numerous human teleoperators, which proves to be costly. 
Given these considerations, it is both crucial and timely to explore this question: how to \textit{scaling up demonstrations in a data-efficient manner} for learning generalist robotic agents?

\begin{figure}[t]
    \centering
    \vspace{-0.4 cm}
    \includegraphics[width=0.95\linewidth]{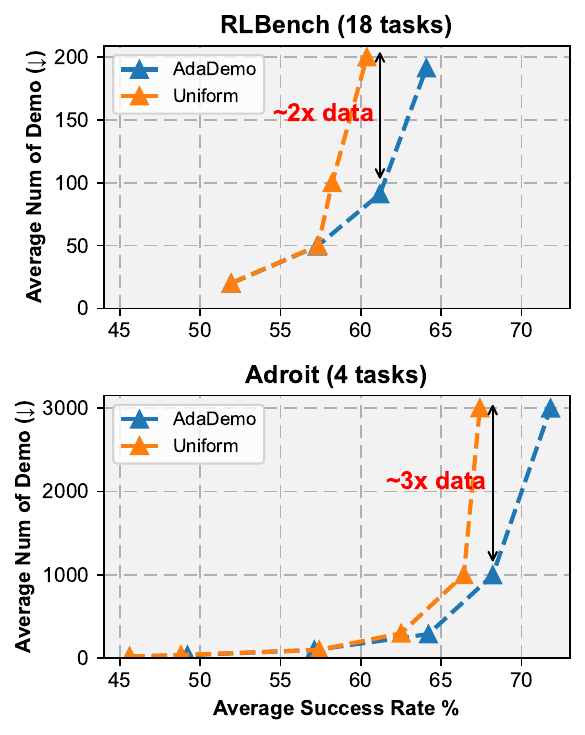}
    \vspace{-0.3 cm}
    \caption{
    Comparison of data efficiency between \textbf{AdaDemo} (adaptively expanding the demo dataset) and \textbf{Uniform} (collecting more demonstrations uniformly). After achieving a mediocre success rate 57\% on RLBench and 62\% on Adroit, \textbf{Uniform} only gains slightly better success rate with a huge increase in demonstration numbers. While the baseline's performance plateaus, \textbf{AdaDemo} continues improving multi-task performance iteratively. Overall, it achieves better performance with only 1/2 the data on RLBench and 1/3 on Adroit. This data efficiency could translate into \textit{substantial cost savings} in large-scale demonstration collection.
    }
    \label{fig:brief_results}
    \vspace{-0.6 cm}
\end{figure}

In this work, we delve into the direction of \textit{actively expanding the demonstration dataset}. Imitation learning \cite{pomerleau1988alvinn} is a widely-used method in training generalist robotic agents, which typically relies on datasets that are \textit{pre-collected and static}. 
Despite various efforts to improve the demonstration datasets through data augmentation \cite{krizhevsky2012imagenet, he2019rethinking} and dataset re-distribution methods \cite{bronstein2023embedding, jiang2021prioritized}, such approaches remain fundamentally limited by the original dataset. 
Our study introduces a significant \textit{paradigm shift}: rather than collecting the demonstration dataset once and using it forever, we propose \textit{actively and continuously expanding the dataset by collecting new demonstrations} to boost the performance of the learned agent. 
In the context of embodied AI, it is usually feasible to collect new demonstrations by interacting with the environments, especially in simulated environments, where the cost of interactions is relatively low.

We study the problem of data-efficient demonstration expansion in a multi-task visual policy learning setup, where the objective is to develop a single generalist policy capable of executing a variety of tasks. This setup aligns with the recent trend of creating versatile, generalist agents that can handle a broad spectrum of challenges \cite{ahn2022can, brohan2022rt, brohan2023rt}. 
A straightforward idea to expand the demonstration dataset is to uniformly collect more demonstrations across all tasks. However, our observations indicate that while this approach can improve policy performance in certain scenarios, it lacks data efficiency and tends to reach a performance plateau easily (see Fig.~\ref{fig:brief_results}).
To improve the data efficiency in demonstration expansion, a crucial intuition is that the \textit{new demonstrations should target scenarios where the current policy fails}.
Stemming from this intuition, we propose a framework named \textbf{AdaDemo} (\textbf{Ada}ptive Online \textbf{Demo}nstration Expansion) that is grounded in three core principles: 1) prioritizing the collection of demonstrations for tasks with low policy performances; 2) within each task, focusing on acquiring demonstrations for the initial states where the policy underperforms; and 3) adapting sampling
strategies in training to emphasize challenging tasks. 
By adopting this framework, we aim to tailor the demonstration expansion process to precisely address the weaknesses of the existing policy, thereby avoiding collecting unnecessary demonstrations.

The effectiveness of AdaDemo was evaluated through a series of multi-task visual policy learning experiments on two robotic manipulation benchmarks: RLBench \cite{rlbench} and Adroit \cite{dapg}, including a total of 22 tasks. These experiments demonstrate AdaDemo's capability to progressively improve policy performance by guiding the generation of high-quality demonstration datasets in a \textit{data-efficient} manner, as depicted in Fig. \ref{fig:brief_results}. 
Notably, AdaDemo exhibits significant data efficiency, \textit{particularly at points where performance plateaus}. Compared to the baseline, it achieves better performance with only 1/2 the data on RLBench and 1/3 on Adroit. This efficiency \textbf{could translate into substantial cost savings}, especially in large-scale demonstration collection.

\noindent To summarize, our contributions go as follows: 
\setlist*[enumerate]{labelindent=\parindent,itemindent=0pt,leftmargin=10pt}
\begin{enumerate}
    \item We propose \textbf{AdaDemo}, a \textbf{data-efficient demonstration expansion framework}, designed to improve multi-task visual policy learning by adaptively expanding the demonstration dataset and employing an appropriate sampling strategy for training.
    \item Through extensive \textbf{experiments on two robotic manipulation benchmarks}, including a total of \textbf{22 tasks}, we demonstrate the effectiveness of AdaDemo in expanding demonstration datasets in a data-efficient way.
\end{enumerate}

\section{Related Work}

\noindent\textbf{Data Augmentation in Policy Learning}
Data augmentation is a widely utilized technique in machine learning that derives additional data samples by applying varied transformations to the existing data. In the context of policy learning, specialized techniques have been proposed: prior works \cite{laskin2020reinforcement, kostrikov2020image, yarats2021mastering} discuss enhancing the robustness of visual policies by applying image-based data augmentation techniques to observations. On the other hand, some recent studies \cite{mandlekar2023mimicgen, pitis2022mocoda} propose to add additional trajectories by adapting existing demonstration trajectories to novel situations. While data augmentation can produce novel data instances, these instances are intrinsically limited in informational value as they are derived from the original dataset.
AdaDemo distinguishes itself from these approaches by directly collecting new demonstrations to address the weakness of the learned policy.

\vspace{0.1 cm}
\noindent\textbf{Data Collection for Foundation Models}
Foundation models, such as large language models \cite{brown2020language}, vision language models \cite{clip}, generalist agents \cite{gato}, and robotic agents \cite{brohan2022rt, brohan2023rt}, usually rely on datasets that are pre-collected and static. This means the data collection phase is conducted once, with the dataset then used forever. Although straightforward to implement, this one-off approach to data collection can restrict the models' potential for learning and adaptation. 
Recently, there has been a shift towards training foundation models using dynamically expanding datasets. For instance, SAM \cite{kirillov2023segment} employs a bootstrapping approach, leveraging pre-trained segmentation models to help annotate a more extensive dataset, which is then used to train stronger models. Similarly, in the robotics field, RoboCat \cite{bousmalis2023robocat} utilizes a pre-trained policy to collect additional demonstrations for new tasks or robots, enlarging the dataset for subsequent agent training.
Compared to these methods, AdaDemo prioritizes data efficiency by selectively targeting data collection efforts towards scenarios where the current policy underperforms, rather than indiscriminately adding data.

\vspace{0.1 cm}
\noindent\textbf{Online Policy Learning }
In contexts where online interaction is feasible, a straightforward strategy for policy learning is using reinforcement learning (RL) \cite{sac}. However, due to the sample efficiency issue and the unstable training dynamics, RL is rarely used in training large-scale policies. 
Also, online imitation learning \cite{dagger, yan2021explaining, chen2022system, liu2019state, mu2024when} presents a viable approach for leveraging online interactions in policy learning. Many of these methods are based on DAgger \cite{dagger}, which requires an expert to provide action supervision for all encountered states during online interactions. The requirement for an expert capable of providing immediate supervision across \textit{all possible states} can be impractical, making such methods less feasible for a wide range of applications. 
In contrast, AdaDemo simplifies this requirement by only requiring a demonstration collector for the \textit{initial states} of tasks. This more manageable assumption positions AdaDemo as a less demanding alternative compared to methods like DAgger.

\section{Problem Setup}

In this paper, we study a multi-task setting where a single visual policy (generalist) is trained to solve multiple tasks, each having variants in terms of different goals and initial states. This setup mirrors practical scenarios, such as a robotic arm performing diverse tasks like stacking plates and opening drawers, depending on the given instructions and observed objects.
We assume that tasks come with a success metric, allowing us to verify the task completion and compute the success rate of the policy. %

To train the multi-task visual policy, we employ %
behavior cloning \cite{pomerleau1988alvinn}, a widely adopted algorithm in learning generalist robotic agents \cite{bousmalis2023robocat, brohan2022rt}. 
Behavior cloning serves as a general foundation that can accommodate various modern visual policy learning architectures \cite{peract, rvt}.
Following prior work, we assume the demonstration dataset in behavior cloning has the format: 
$\data_0 := \{ \data^1_0, ..., \data^{M}_0 \},$
where $M$ is the number of tasks, the superscript denotes the index of the task, and the subscript $0$ denotes this is an initial pre-collected demonstration dataset.
The demonstration dataset of each task $m$ contains a set of demonstration trajectories $\tau$:
$\data^m_0 := \{\tau^m_0, \tau^m_1, ...  \}$.
For each demonstration $\tau$, it contains a goal description $g$ and a sequence of transition: $\tau := \langle g, \{(o_0, a_0), (o_1, a_1), ...\}\rangle,$
where $g$ is a goal description (e.g., language or low-dim vectors) and $o_i$ is a visual observation,
, which serves as the input of the visual policy.
$a_i$ is an action, which provides the supervision for output of visual policy. %

An important aspect of our problem setup is the \textbf{flexibility in collecting demonstration}, which can be sourced through varied methods such as task and motion planning, state-based reinforcement learning, or teleoperation from human demonstrators. 
We assume access to a demonstration collector to \textit{generate demonstrations for any state from the initial state distribution of the tasks}. Note that the demonstration collector \textbf{only needs to handle the initial states} of tasks, thus the assumption is less demanding than methods like DAgger \cite{dagger}, which require an oracle providing action supervision across \textbf{all possible states}. Such a demonstration collector is also feasible in real world via human teleoperation \cite{fu2024mobile} or real-world motion planning \cite{danielczuk2021object, fishman2023motion}.

Generating demonstrations for all possible states is significantly more challenging than focusing on initial states, because most human-designed tasks are typically solvable from the initial states. For instance, in a "Stacking Cup" task, it is feasible to generate demonstrations where cups are initially standing on a table. However, if a cup is knocked down or falls from the table during the task execution, creating a demonstration for that state becomes difficult or impossible. Thus, our assumption is more realistic, only requiring a demonstration collector capable of handling initial task states.

\section{Adaptive Online Demo Expansion}

\subsection{Overview}

We aim to \textit{actively} and \textit{continuously} expand the demonstration dataset for training generalist robotic agents, particularly, for multi-task visual policy learning. 
Contrary to the conventional use of pre-collected and static datasets in imitation learning \cite{pomerleau1988alvinn}, \textbf{AdaDemo} (\textbf{Ada}ptive Online \textbf{Demo}nstration Expansion), differentiates itself by adopting an online and adaptive method. It operates over multiple iterative rounds to progressively improve the demonstration dataset, and consequently, the performance of the policy. 
Before delving into the technical details, we first highlight the key properties of AdaDemo below:

\subsubsection{Online Demonstration Expansion} 

AdaDemo advocates the concept of online expansion of the demonstration dataset to overcome the limitations associated with pre-collected and static datasets in imitation learning. 
By incorporating new demonstrations, it becomes possible to address the weakness of the learned policy by directly providing additional supervision. 
This strategy is also supported by recent successes in foundation models \cite{kirillov2023segment, bousmalis2023robocat}, suggesting that expanding the dataset is a potent method for enhancing model performance.

\subsubsection{Iterative Improvement Process}
AdaDemo employs an iterative approach, progressively incorporating new demonstrations into the dataset with each round. This iterative process offers several key advantages over the bulk accumulation of demonstrations:
\textbf{1) Data efficiency:}
With the policy re-evaluated at each round, it is possible to specifically collect new demonstrations that address the current policy's weakness, thus eliminating the collection of unnecessary demonstrations;
\textbf{2) Adaptability to Dynamic Budgets:}
Resource budgets for demonstration collection may vary over time. This iterative approach accommodates such fluctuations, allowing for the initial use of available resources and the integration of new demonstrations as more resources become available;
\textbf{3) Alignment with Current Research:}
The iterative expansion of data is a strategy that has been utilized in modern foundation models \cite{kirillov2023segment}, underscoring its effectiveness and relevance to current practices in the field.

\begin{figure}[t]
    \centering
    \vspace{0.2 cm}
    \includegraphics[width=\linewidth]{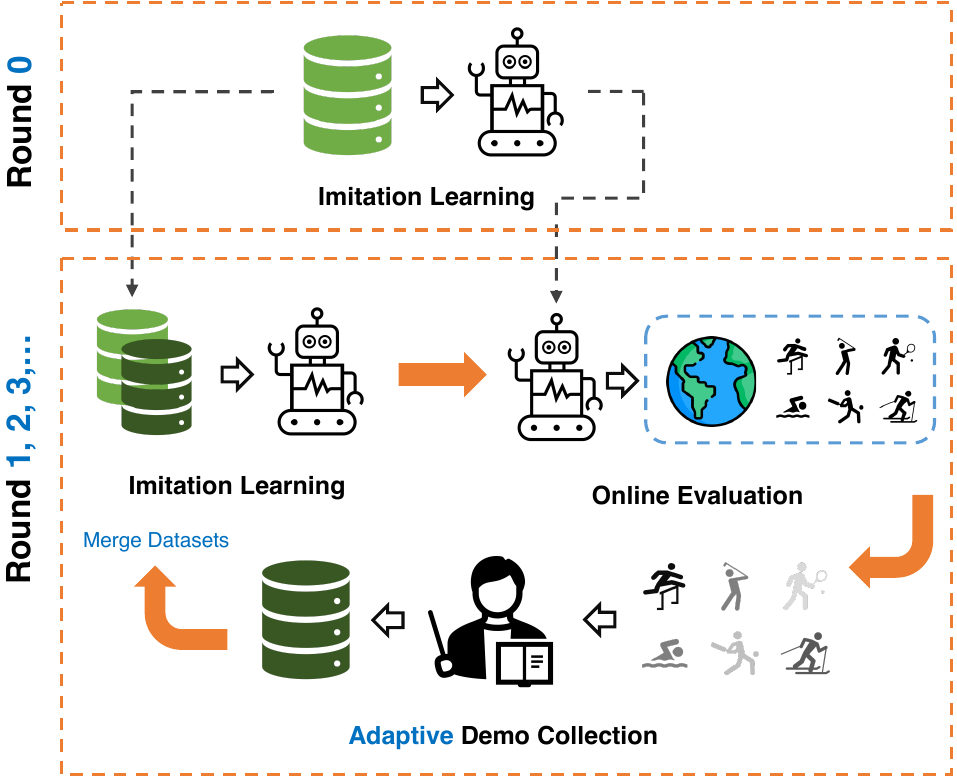}
    \caption{
    \textbf{AdaDemo} iteratively expands the demonstration dataset through online evaluation of the trained policy, adaptively collecting additional demonstrations to target cases where the multi-task policy most needs improvement.
    }
    \vspace{-0.6 cm}
    \label{fig:method_overview}
\end{figure}

\subsubsection{Adaptive Demonstration Expansion} 
\label{sec:adpative_expansion}

Instead of indiscriminately adding data, AdaDemo strategically focuses on demonstrations that are likely to yield the most significant improvements in policy performance, thus ensuring enhanced data efficiency.
The adaptiveness is mainly grounded in \textbf{three core principles}:
1) Emphasizing collecting more demonstrations for initial states where the current policy fails (Sec. \ref{sec:demo_on_failed});
2) Prioritizing the collection of demonstrations for tasks where the current policy gets low success rates (Sec. \ref{sec:demo_on_hard});
3) to efficiently utilize the additional data, AdaDemo adapts the sampling strategy in training to emphasize tasks considered more challenging, ensuring that the model is continually exposed to and learns from the most demanding scenarios (Sec. \ref{sec:sample_startegy}).

In summary, AdaDemo actively expands the demonstration dataset for training generalist robotic agents. It emphasizes an online, iterative, and adaptive strategy to expand the dataset, ensuring that each newly incorporated demonstration is strategically selected to boost the policy's capability across a diverse range of tasks. 
See Fig. \ref{fig:method_overview} for a visual illustration of our framework. 
The full framework is summarized in Algorithm \ref{alg:ours}.
In the following subsections, we delve into the details of the three core principles mentioned in Sec. \ref{sec:adpative_expansion}.

\subsection{Demo Collection on Failed Initial States}
\label{sec:demo_on_failed}

One core principle of AdaDemo is to focus on collecting demonstrations for tasks specifically from those \textit{initial states} where the current policy falters. To systematically identify these failed initial states, we evaluate the multi-task policy on each task, pinpointing specific initial states where the current policy cannot successfully complete the task. 
For the sake of brevity, we treat the goal description as a part of the initial state of a task. 
Focusing on these failed initial states ensures that the newly collected demonstrations provide direct guidance for solving scenarios where the policy previously failed, thereby enhancing the data efficiency without wasting resources on collecting demonstrations for initial states the policy already navigates successfully.

In practice, we evaluate the policy on the default initial state distribution for each task. Upon encountering an initial state that leads to an unsuccessful episode, we engage a demonstration collector to generate an expert demonstration trajectory from that specific initial state.

Our framework is compatible with a variety of tools for expert demonstration collection, including task and motion planning systems, state-based reinforcement learning, model-predictive control, and even human teleoperation. 
The key requirement for these tools is their capability to complete the task from a given initial state. In cases where a demonstration collector fails to complete a task from a particular initial state, we can either opt for retrying the collection process a few more times, or choose to skip collecting a demonstration for that specific initial state.
This flexible approach accommodates the reality of imperfect demonstration collectors, ensuring our framework's broad applicability.

\subsection{Demo Collection on Unsolved Tasks}
\label{sec:demo_on_hard}

With the above strategy for collecting new demonstrations within a task, a critical consideration emerges: how should we allocate our demonstration collection budget across various tasks? Specifically, we need to decide the number of demonstrations to collect for each task.

Our guiding principle is straightforward: prioritize the collection of demonstrations for tasks that remain unsolved, i.e., the tasks where the current policy gets low success rates. These tasks are crucial in improving the overall performance across all tasks, because they offer large room for improvement. 
For each task, we keep evaluating the policy until it reaches a predetermined number of successful episodes. During this process, demonstrations are collected from the initial states where the policy fails, as described in Sec. \ref{sec:demo_on_failed}.

Consequently, tasks with lower success rates naturally lead to more failures during the evaluation, thereby getting more new demonstrations. This approach creates a direct correlation between the task's difficulty (gauged by its success rate) and the number of demonstrations collected for this task. It ensures a concentrated effort on those tasks where the current policy underperforms, leading to a more efficient and focused data-collection process.

To provide a quantitative analysis, the formula for estimating the number of demonstrations $N_{demo}^k$ required for each task $k$ is given by $N_{demo}^k = \frac{E}{SR^k}-E,$, where $SR^k$ denotes the success rate of the current policy on task $k$, and $E$ is a hyperparameter specifying the target number of success episodes, which could be adjusted based on the available resource budget. 
In practice, we also impose a cap on the demonstrations to be collected per task. The rationale behind this is to prevent the potential endless evaluation process to reach a target number of successful episodes for some tasks with extremely low success rates.

\setlength{\textfloatsep}{5pt}
\begin{algorithm}[t]
    \caption{Adpative Online Demo Expansion}
    \small
    \label{alg:ours}
    \begin{algorithmic}[1]
        \Require Initial demonstration dataset for $M$ tasks $\data_0 := \{\data^0_0, \data^1_0, ..., \data^{M}_0 \}$, where demonstration dataset for task $k$ $\data^k_0 := \{\tau^k_0, \tau^k_1, ...  \}$; Demonstration collector $G$;
    
        \State Train a multi-task policy $\policy$ on $\data_0$ by imitation learning
        \For {each round $i$}
            \State Determine a target number of success episodes $E_i$ for this round according to the budget
            \For {each task $T^k$}
                \State $\data_i^k\leftarrow \{\}$
                \State Counter $c\leftarrow 0$
                \Repeat
                    \State Random sample an initial state $s$ in task $T^k$
                    \State Evaluate $\policy$ on $s$
                    \If {$\policy$ failed on $s$}
                        \State Collect a demo trajectory on $s$: $\tau \leftarrow G(s)$
                        \State $\data_i^k \leftarrow \data_i^k \cup \{\tau \}$
                    \Else
                        \State $c\leftarrow c + 1$
                    \EndIf
                \Until {$c = E_i$ }
            \EndFor
            \State $\data_i \leftarrow \{\data^0_i, \data^1_i, ..., \data^{M}_i \}$
            \State Merge the datasets from all rounds $\data \leftarrow \bigcup_{j=0}^i \data_i$
            \State Re-train $\policy$ on $\data$ with the selected sampling strategy
    
        \EndFor
    \end{algorithmic}
\end{algorithm}

\subsection{Sampling Strategy in the Collected Dataset}
\label{sec:sample_startegy}

Beyond increasing the number of demonstrations, it is also crucial to optimize the use of the expanded dataset through a suitable sampling strategy during training. A simplistic strategy might involve uniformly sampling across the entire dataset. However, this method unintentionally biases towards tasks with longer trajectories, which does not necessarily correlate with their actual importance to the overall policy performance. To address this bias, we need to adopt alternative sampling strategies that more accurately prioritize demonstrations which are crucial for enhancing policy performance.
One possible approach could be to ensure uniform sampling with respect to tasks, granting each task an equal chance of being selected. This method offers a more balanced distribution compared to uniform sampling with respect to the entire dataset, but falls short in highlighting the significance of the more difficult tasks, which are vital for the overall policy performance.

Therefore, we adopt a strategy that draws samples uniformly with respect to the demonstration trajectories, i.e., the sampling probability of each task's data is proportional to the number of demonstration trajectories in the dataset. This method aligns with our focused approach of collecting more demonstrations for unsolved tasks. By sampling in this manner, we naturally place more emphasis on these unsolved tasks within the learning process. This not only ensures that harder tasks receive more attention, but also maximizes the utility of the additional demonstrations we have collected. In practice, we also set a minimum sampling weight for each task to avoid some tasks with a very small number of demonstrations being undersampled during training.

This sampling strategy plays a crucial role in our framework, guaranteeing that the expanded demonstration dataset is fully utilized, thereby directly supporting our objective of achieving maximal data efficiency.

\begin{figure*}[t]
    \vspace{0.2 cm}
    \begin{minipage}{0.28\textwidth}
        \begin{subfigure}{\textwidth}
            \centering
            \includegraphics[width=0.42\textwidth]{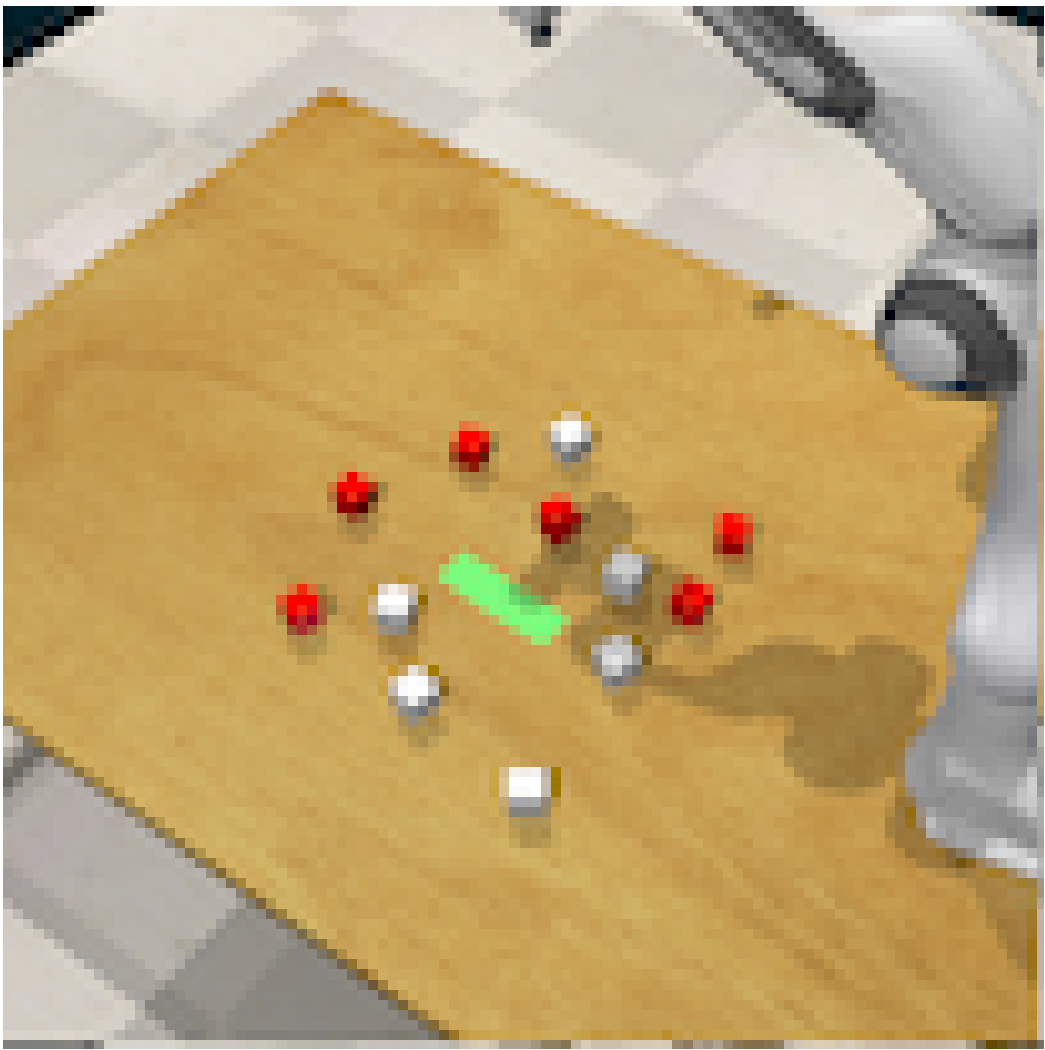}
            \includegraphics[width=0.42\textwidth]{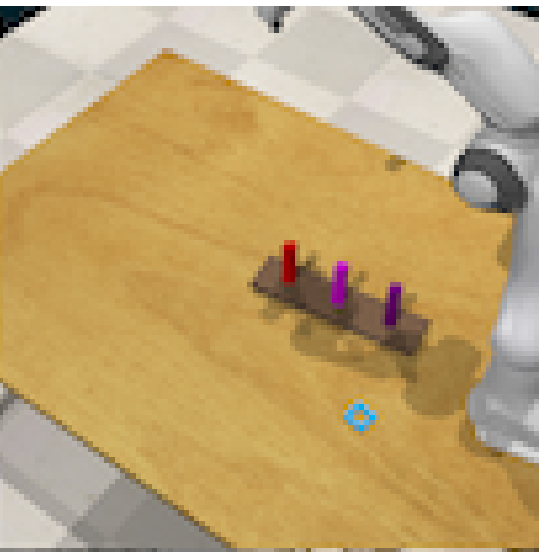}
            \centering
            \\
            \vspace{0.1 cm}
            \includegraphics[width=0.45\textwidth]{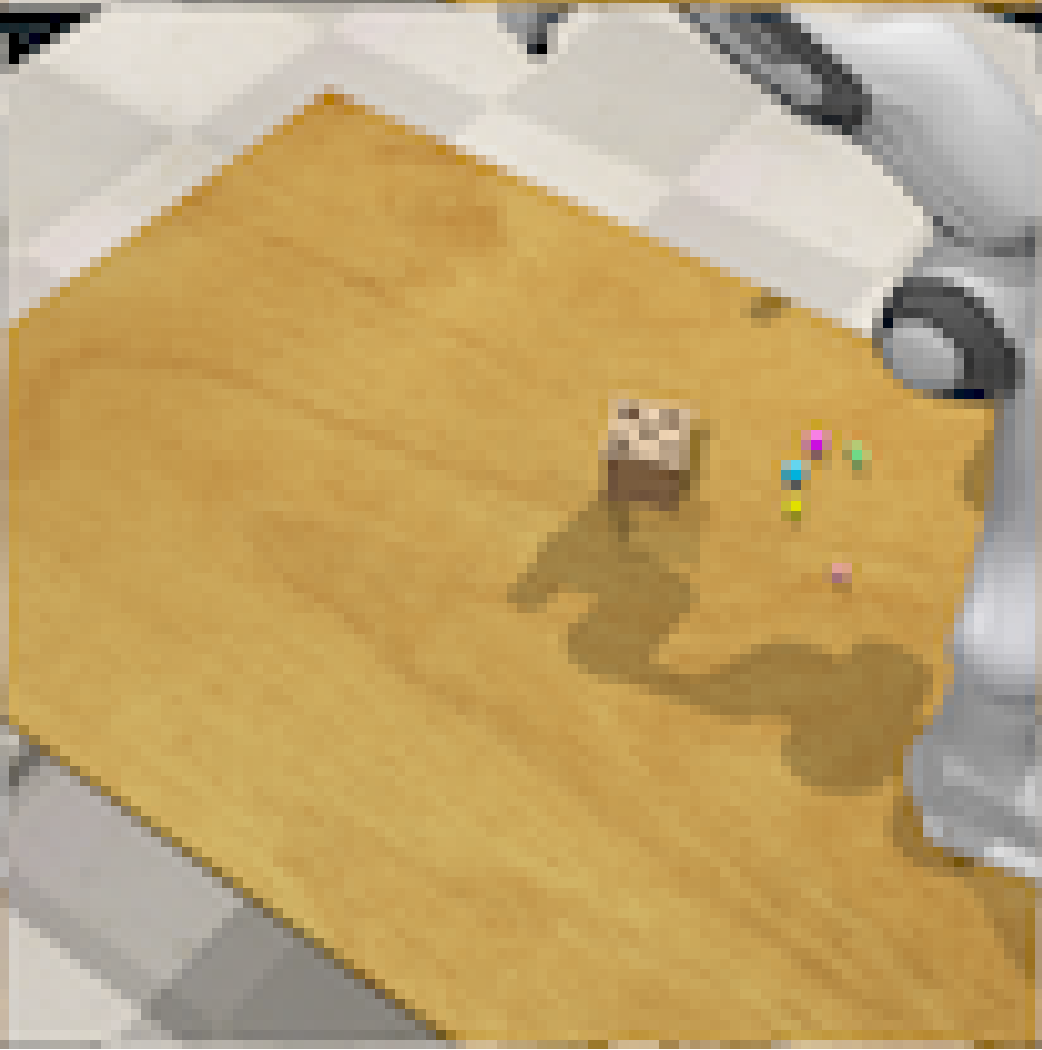}
            \includegraphics[width=0.42\textwidth]{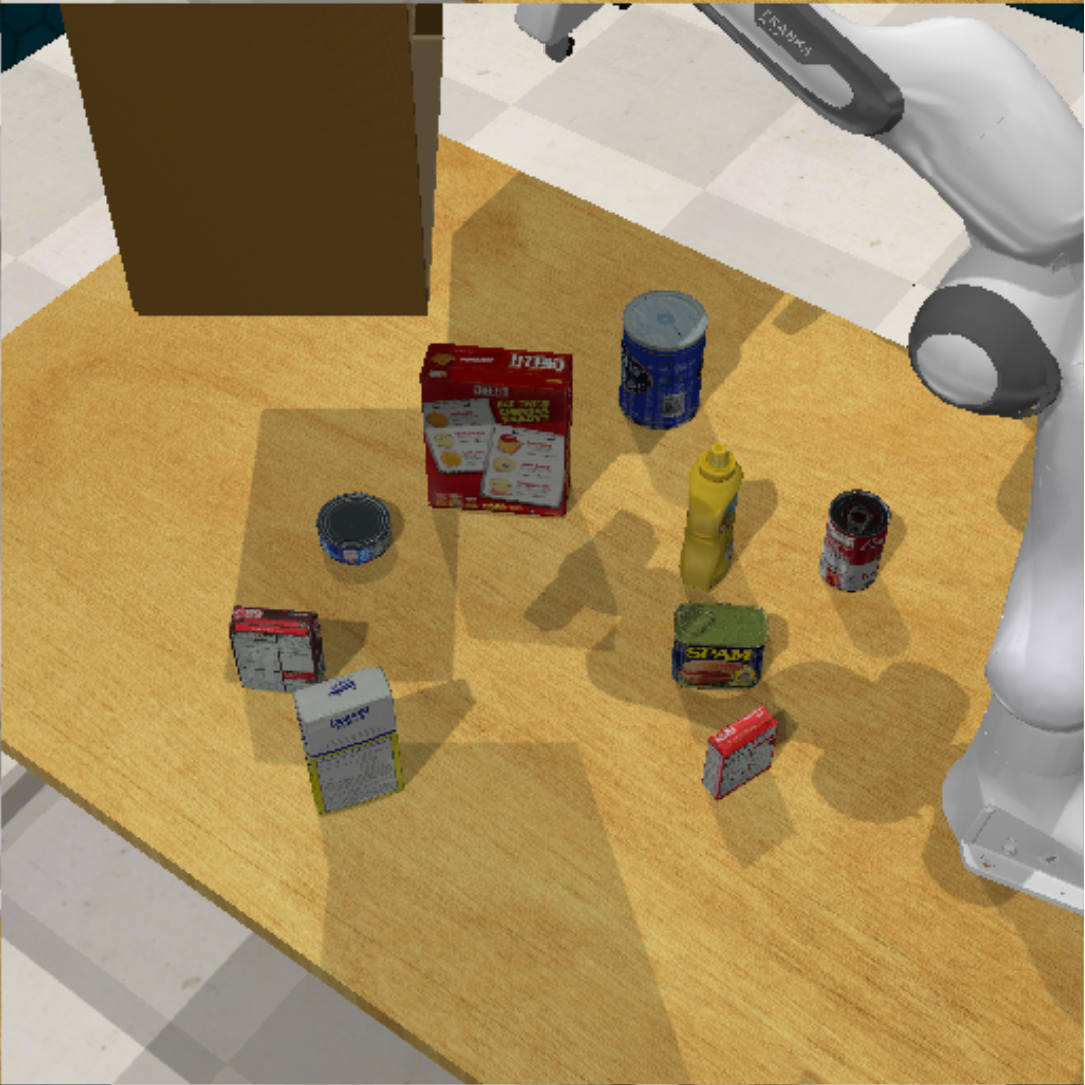}
            \caption{RLBench (showing 4 of 18 tasks)}
        \end{subfigure}
        \begin{subfigure}{\textwidth}
            \centering
            \vspace{0.1 cm}
            \includegraphics[width=0.42\textwidth]{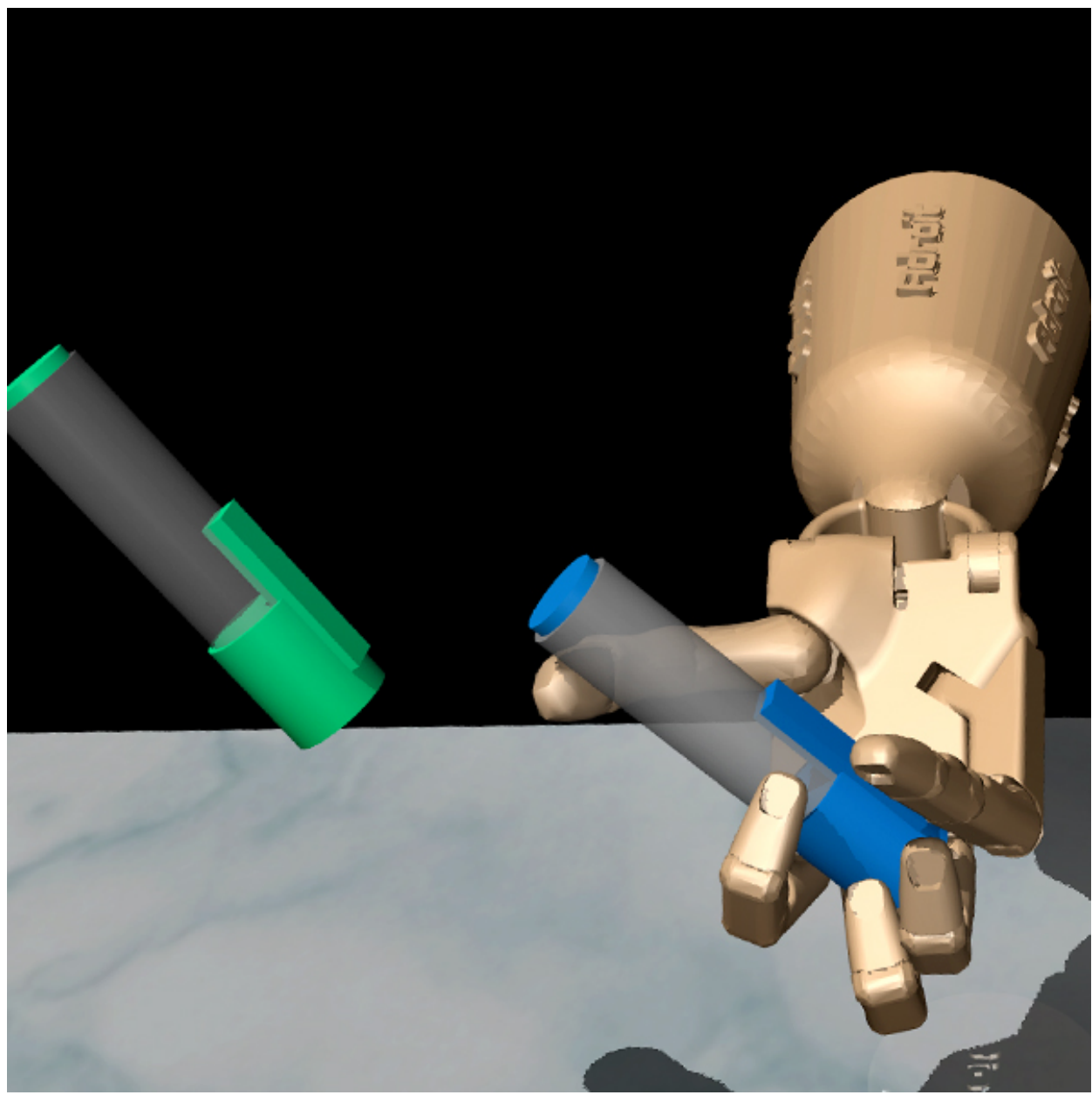}
            \includegraphics[width=0.42\textwidth]{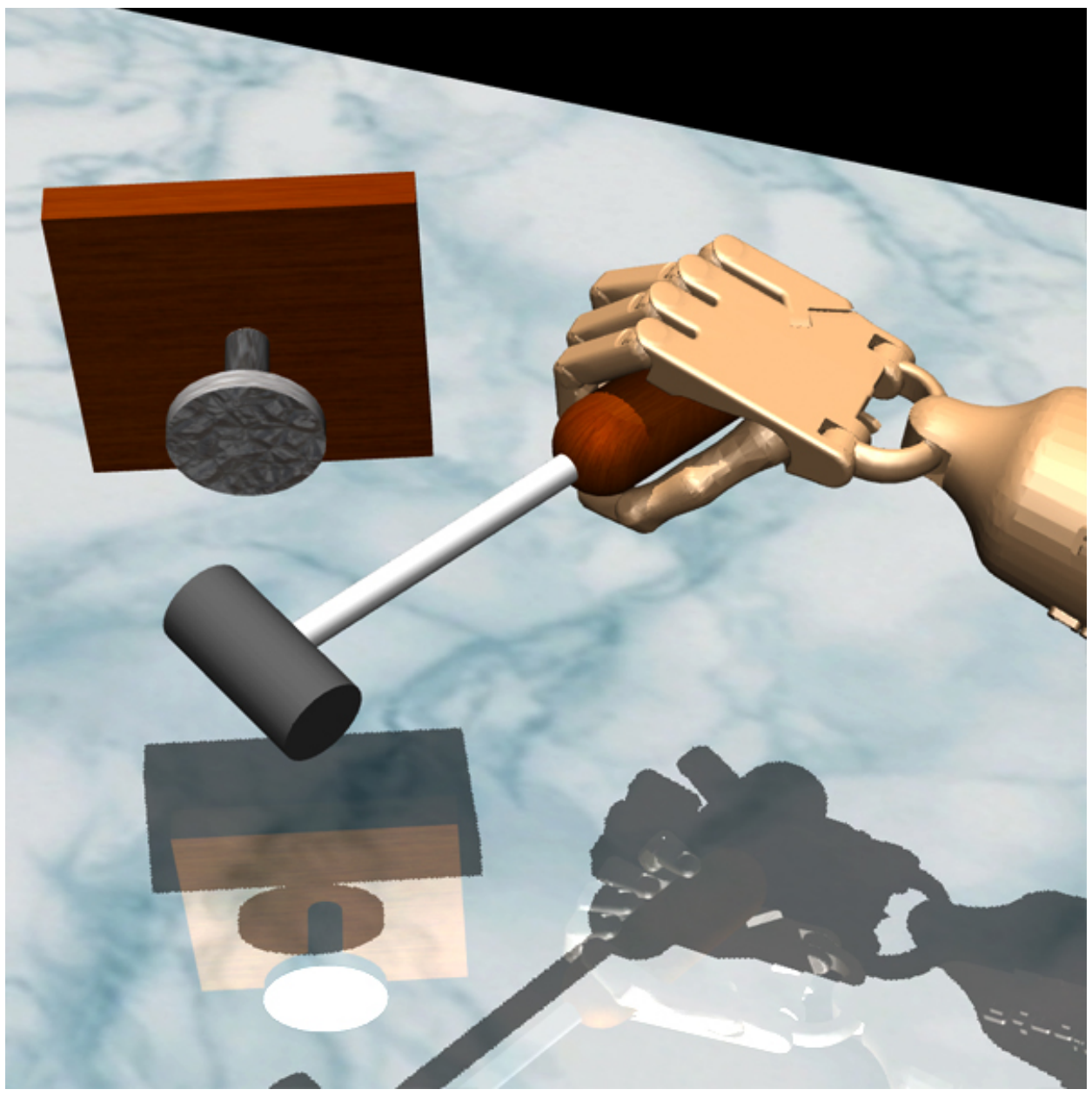}
            \caption{Adriot (showing 2 of 4 tasks)}
        \end{subfigure}
        \captionof{figure}{\textbf{Tasks:} We consider challenging and diverse robotic manipulation tasks spanning two benchmarks: RLBench (table-top robot arm manipulation) and Adroit (dexterous manipulation).}
        \label{fig:task_example}
    \end{minipage}
    \hfill
    \begin{minipage}{0.7\textwidth}
        \centering
\resizebox{\columnwidth}{!}{
\setlength\tabcolsep{2pt} %
\begin{tabular}{c|c|cc|cc|cc}
\toprule
\rowcolor[HTML]{D7DAFF}
                 & Initial & \multicolumn{2}{c|}{Round 1} & \multicolumn{2}{c|}{Round 2} & \multicolumn{2}{c}{Round 3} \\
                 &         & \textcolor{blue}{$\bullet$} AdaDemo         & \textcolor{orange}{$\bullet$} Uniform        & \textcolor{blue}{$\bullet$} AdaDemo         & \textcolor{orange}{$\bullet$} Uniform        & \textcolor{blue}{$\bullet$} AdaDemo         & \textcolor{orange}{$\bullet$} Uniform        \\
\midrule
Put in Drawer    & 85.3{\footnotesize ±8.3}          & 80.0{\footnotesize ±4.0}  & 80.0{\footnotesize ±17.4}         & 72.0{\footnotesize ±12.0}         & 70.7{\footnotesize ±31.1}         & 82.7{\footnotesize ±12.2}          & \textbf{94.7}{\footnotesize ±2.3}  \\
\rowcolor[HTML]{EFEFEF}
Drag Stick       & 78.7{\footnotesize ±15.1}         & 78.7{\footnotesize ±16.2} & 82.7{\footnotesize ±23.1}         & \textbf{97.3}{\footnotesize ±2.3} & 65.3{\footnotesize ±22.7}         & 92.0{\footnotesize ±4.0}           & 58.7{\footnotesize ±40.3}          \\
Turn Tap         & 92.0{\footnotesize ±4.0}          & 90.7{\footnotesize ±2.3}  & 96.0{\footnotesize ±4.0}          & 92.0{\footnotesize ±10.6}         & 93.3{\footnotesize ±2.3}          & \textbf{96.0}{\footnotesize ±4.0}  & 93.3{\footnotesize ±2.3}           \\
\rowcolor[HTML]{EFEFEF}
Slide Block      & \textbf{56.0}{\footnotesize ±8.0} & 33.3{\footnotesize ±8.3}  & 54.7{\footnotesize ±22.0}         & 16.0{\footnotesize ±4.0}          & 29.3{\footnotesize ±18.0}         & 17.3{\footnotesize ±6.1}           & 13.3{\footnotesize ±8.3}           \\
Open Drawer      & 78.7{\footnotesize ±8.3}          & 85.3{\footnotesize ±8.3}  & 76.0{\footnotesize ±10.6}         & \textbf{85.3}{\footnotesize ±6.1} & 81.3{\footnotesize ±6.1}          & 80.0{\footnotesize ±4.0}           & 74.7{\footnotesize ±4.6}           \\
\rowcolor[HTML]{EFEFEF}
Put in Cupboard  & 34.7{\footnotesize ±8.3}          & 61.3{\footnotesize ±4.6}  & 58.7{\footnotesize ±8.3}          & 64.0{\footnotesize ±6.9}          & 65.3{\footnotesize ±2.3}          & \textbf{77.3}{\footnotesize ±12.9} & 60.0{\footnotesize ±13.9}          \\
Sort Shape       & 29.3{\footnotesize ±2.3}          & 28.0{\footnotesize ±4.0}  & 36.0{\footnotesize ±8.0}          & 40.0{\footnotesize ±4.0}          & 40.0{\footnotesize ±4.0}          & \textbf{41.3}{\footnotesize ±10.1} & 37.3{\footnotesize ±8.3}           \\
\rowcolor[HTML]{EFEFEF}
Put in Safe      & 76.0{\footnotesize ±4.0}          & 78.7{\footnotesize ±8.3}  & 90.7{\footnotesize ±2.3}          & 86.7{\footnotesize ±6.1}          & \textbf{93.3}{\footnotesize ±2.3} & \textbf{93.3{\footnotesize ±2.3}}  & 93.3{\footnotesize ±4.6}           \\
Push Buttons     & 74.7{\footnotesize ±4.6}          & 86.7{\footnotesize ±2.3}  & 88.0{\footnotesize ±0.0}          & 92.0{\footnotesize ±4.0}          & 96.0{\footnotesize ±4.0}          & 96.0{\footnotesize ±4.0}           & \textbf{98.7}{\footnotesize ±2.3}  \\
\rowcolor[HTML]{EFEFEF}
Close Jar        & 29.3{\footnotesize ±6.1}          & 45.3{\footnotesize ±2.3}  & 38.7{\footnotesize ±12.2}         & \textbf{58.7}{\footnotesize ±4.6} & 48.0{\footnotesize ±10.6}         & 48.0{\footnotesize ±6.9}           & 48.0{\footnotesize ±13.9}          \\
Stack Blocks     & 13.3{\footnotesize ±8.3}          & 29.3{\footnotesize ±10.1} & 17.3{\footnotesize ±12.9}         & 37.3{\footnotesize ±9.2}          & 25.3{\footnotesize ±2.3}          & \textbf{46.7}{\footnotesize ±2.3}  & 37.3{\footnotesize ±8.3}           \\
\rowcolor[HTML]{EFEFEF}
Place Cups       & 1.3{\footnotesize ±2.3}           & 0.0{\footnotesize ±0.0}   & 0.0{\footnotesize ±0.0}           & 2.7{\footnotesize ±2.3}           & \textbf{8.0}{\footnotesize ±0.0}  & 1.3{\footnotesize ±2.3}            & 4.0{\footnotesize ±4.0}            \\
Place Wine       & 92.0{\footnotesize ±4.0}          & 84.0{\footnotesize ±8.0}  & \textbf{94.7}{\footnotesize ±2.3} & 78.7{\footnotesize ±4.6}          & 85.3{\footnotesize ±2.3}          & 90.7{\footnotesize ±6.1}           & 92.0{\footnotesize ±4.0}           \\
\rowcolor[HTML]{EFEFEF}
Screw Bulb       & 38.7{\footnotesize ±4.6}          & 50.7{\footnotesize ±8.3}  & 49.3{\footnotesize ±4.6}          & 57.3{\footnotesize ±2.3}          & 58.7{\footnotesize ±18.0}         & \textbf{78.7}{\footnotesize ±4.6}  & 62.7{\footnotesize ±8.3}           \\
Sweep to Dustpan & 45.3{\footnotesize ±11.5}         & 52.0{\footnotesize ±4.0}  & 48.0{\footnotesize ±6.9}          & 48.0{\footnotesize ±12.0}         & 46.7{\footnotesize ±11.5}         & 45.3{\footnotesize ±2.3}           & \textbf{57.3}{\footnotesize ±4.6}  \\
\rowcolor[HTML]{EFEFEF}
Insert Peg       & 9.3{\footnotesize ±4.6}           & 13.3{\footnotesize ±6.1}  & 4.0{\footnotesize ±4.0}           & \textbf{21.3}{\footnotesize ±4.6} & 16.0{\footnotesize ±8.0}          & 12.0{\footnotesize ±0.0}           & 16.0{\footnotesize ±4.0}           \\
Meat off Grill   & 96.0{\footnotesize ±4.0}          & 97.3{\footnotesize ±2.3}  & 96.0{\footnotesize ±0.0}          & 97.3{\footnotesize ±2.3}          & 96.0{\footnotesize ±0.0}          & 94.7{\footnotesize ±4.6}           & \textbf{100.0}{\footnotesize ±0.0} \\
\rowcolor[HTML]{EFEFEF}
Stack Cups       & 4.0{\footnotesize ±4.0}           & 37.3{\footnotesize ±8.3}  & 20.0{\footnotesize ±21.2}         & 54.7{\footnotesize ±6.1}          & 29.3{\footnotesize ±6.1}          & \textbf{60.0}{\footnotesize ±6.9}  & 45.3{\footnotesize ±20.5}          \\
\midrule
Average SR       & 51.9              & 57.3      & 57.3              & 61.2              & 58.2              & \textbf{64.1}      & 60.4               \\
\rowcolor[HTML]{EFEFEF}
Average \# Demo  & 20                & 49.4      & 50                & 91.0              & 100               & 191.2              & 200                \\
\bottomrule
\end{tabular}

}

        \vspace{-0.2 cm}
        \captionof{table}{
            \textbf{RLBench Results:} "SR" stands for success rate, and "Average \# Demo" indicates the average number of demonstrations allocated per task. "Uniform" refers to the baseline where demonstrations are uniformly collected across all tasks and initial states. SRs are averaged over 3 random seeds. All agents are evaluated for 100 episodes on each task.
        }
        \label{tab:rlbench_result}
    \end{minipage}
\end{figure*}

\begin{table*}[t]
\vspace{-0.1 cm}
\centering
\setlength\tabcolsep{2pt} %
\begin{tabular}{c|c|cc|cc|cc|cc|cc}
\toprule
\rowcolor[HTML]{D7DAFF}
                & Initial  & \multicolumn{2}{c|}{Round 1} & \multicolumn{2}{c|}{Round 2} & \multicolumn{2}{c|}{Round 3} & \multicolumn{2}{c|}{Round 4} & \multicolumn{2}{c}{Round 5}  \\
                &          & {\footnotesize \textcolor{blue}{$\bullet$} {\footnotesize AdaDemo}}         & {\footnotesize \textcolor{orange}{$\bullet$} {\footnotesize Uniform}}      & {\footnotesize \textcolor{blue}{$\bullet$} {\footnotesize AdaDemo}}         & {\footnotesize \textcolor{orange}{$\bullet$} {\footnotesize Uniform}}      & {\footnotesize \textcolor{blue}{$\bullet$} {\footnotesize AdaDemo}}         & {\footnotesize \textcolor{orange}{$\bullet$} {\footnotesize Uniform}}      & {\footnotesize \textcolor{blue}{$\bullet$} {\footnotesize AdaDemo}}              & {\footnotesize \textcolor{orange}{$\bullet$} {\footnotesize Uniform}} & {\footnotesize \textcolor{blue}{$\bullet$} {\footnotesize AdaDemo}}              & {\footnotesize \textcolor{orange}{$\bullet$} {\footnotesize Uniform}}  \\
\midrule
Relocate        & 0.6{\footnotesize ±0.4}  & 1.9{\footnotesize ±0.7}      & 0.9{\footnotesize ±0.7}      & 6.0{\footnotesize ±1.7}      & 6.0{\footnotesize ±3.1}      & 18.6{\footnotesize ±4.0}     & 12.3{\footnotesize ±0.9}     & 28.8{\footnotesize ±1.3} & 25.2{\footnotesize ±5.5} & \textbf{32.2}{\footnotesize ±6.2} & 23.6{\footnotesize ±9.6} \\
\rowcolor[HTML]{EFEFEF}
Door            & 50.0{\footnotesize ±3.1} & 55.2{\footnotesize ±8.5}     & 55.7{\footnotesize ±6.6}     & 70.1{\footnotesize ±3.3}     & 70.5{\footnotesize ±3.2}     & 73.7{\footnotesize ±3.5}     & 76.1{\footnotesize ±1.2}     & 77.2{\footnotesize ±3.0} & 79.7{\footnotesize ±3.8} & \textbf{83.0}{\footnotesize ±3.9} & 74.4{\footnotesize ±4.1} \\
Pen             & 35.8{\footnotesize ±1.4} & 42.7{\footnotesize ±1.9}     & 39.8{\footnotesize ±0.7}     & 53.5{\footnotesize ±2.4}     & 55.5{\footnotesize ±5.5}     & 66.6{\footnotesize ±6.4}     & 64.3{\footnotesize ±4.9}     & 68.9{\footnotesize ±1.3} & 63.0{\footnotesize ±3.2} & \textbf{74.8}{\footnotesize ±1.5} & 74.6{\footnotesize ±3.2} \\
\rowcolor[HTML]{EFEFEF}
Hammer          & 95.7{\footnotesize ±4.9} & 97.2{\footnotesize ±1.6}     & \textbf{98.9}{\footnotesize ±1.7}     & 98.8{\footnotesize ±1.0}     & 97.7{\footnotesize ±1.8}     & 97.6{\footnotesize ±1.7}     & 97.5{\footnotesize ±2.4}     & 97.7{\footnotesize ±3.5} & 97.7{\footnotesize ±2.6} & 97.3{\footnotesize ±4.4} & 96.9{\footnotesize ±4.7} \\
\midrule
Average SR      & 45.6     & 49.2         & 48.8         & 57.1         & 57.4         & 64.2         & 62.5         & 68.2     & 66.4  & \textbf{71.8}     & 67.4     \\
\rowcolor[HTML]{EFEFEF}
Average \# Demo & 20       & 37.8         & 40           & 99.0         & 100          & 290.0        & 300          & 997.0    & 1000  & 2994.3     & 3000    \\
\bottomrule
\end{tabular}

\caption{
\textbf{Adroit Results:} For definitions of specific terms, please refer to Table \ref{tab:rlbench_result}.
}
\label{tab:adroit_result}

\vspace{-0.5 cm}
\end{table*}

\section{Experiments}

\subsection{Experimental Setup}

The goal of our experimental evaluation is to study \textit{whether AdaDemo can effectively expand the demonstration dataset in a data-efficient manner, thereby enhancing the learned multi-task visual policy.}
Conceptually, AdaDemo is expected to be compatible with a variety of core components within the learning system, including the demonstration collector, network architecture, and type of controller. To confirm AdaDemo's wide-ranging applicability, our experimental designs incorporate variations across these dimensions:
\begin{itemize}
    \item \textbf{2 Demonstration collectors}: Task and Motion Planning (TAMP),  and state-based reinforcement learning (RL)
    \item \textbf{2 Controllers}: Keyframe-based end-effector control, and joint position control
    \item \textbf{2 Network architectures}: Vision Transformer \cite{rvt}, and Convolutional Neural Network \cite{lfs}.
    \item \textbf{A total of 22 tasks across 2 benchmarks}: RLBench \cite{rlbench} (table-top robotic manipulation, 18 tasks) and Adroit \cite{dapg} (dexterous manipulation, 4 tasks). Fig. \ref{fig:task_example} shows sample tasks from each benchmark.
\end{itemize}

\noindent We summarize the key details of our setups as follows.

\subsubsection{Environments: RLBench}

\paragraph{Task Description} We follow the standard setting used in \cite{peract, rvt}. A 7-DoF Franka Panda robot equipped with a parallel gripper is directed to solve a total of 18 tasks, including pick-and-place, tool use, drawer opening, and high-accuracy peg insertions. Each task includes several variations specified by the associated language description. The visual observations are from four RGB-D cameras positioned at the robot's front side, left shoulder, right shoulder, and wrist, each providing a resolution of 128 × 128. 

\paragraph{Demonstration Collector} The demonstrations are collected by a task and motion planning (TAMP) system from RLBench \cite{rlbench}. Notably, this TAMP system can only solve the tasks from the default task initial states. It lacks the capacity to tackle the tasks from certain intermediate states. 

\paragraph{Visual Policy and Controller} 
In our RLBench experiments, we employ RVT \cite{rvt} as the backbone visual policy. RVT features a multi-view transformer architecture and performs keyframe-based control, predicting actions at low frequency that specify the next keyframe's end-effector pose. A motion planner then directs the end effector to this predicted pose. The original RVT (as well as its predecessor \cite{peract, james2022q}) has a \href{https://github.com/peract/peract/issues/6#issuecomment-1355555980}{legacy issue} leading to that makes the training dataset unnecessarily large. By refining the data processing scripts, we have mitigated this, resulting in a more storage-efficient RVT with a minor performance trade-off. 
Importantly, since the same modifications are consistently applied across all our RLBench experiments, we ensure a fair comparison between AdaDemo and baselines.
Additionally, to accommodate the increasing demonstration data, we also raise the number of epochs to ensure that the model adequately fits the data.

\subsubsection{Environments: Adroit} 

\paragraph{Task Description} Adroit tasks are introduced in \cite{dapg}, where a 24-DoF ADROIT hand is tasked to master four dexterous manipulation skills: object relocation, in-hand manipulation, tool use, and opening doors. The visual observations are acquired from a pre-defined camera with a resolution of 256 × 256. 

\paragraph{Demonstration Collector} The demonstrations are collected by a well-trained state-based RL agent. It is a common practice to employ state-based RL agents for demonstration collection, followed by rendering to images \cite{mu2021maniskill, gu2023maniskill2, wan2023unidexgrasp}. This approach is favored because generally it is easier to train a state-based agent compared to a visual agent.

\paragraph{Visual Policy and Controller}
In our Adroit experiments, we utilize LfS \cite{lfs} as the backbone visual policy. This approach employs a CNN enhanced with random shifting augmentation to generate actions at a high frequency, controlling the absolute joint positions of the dexterous hand. For the sake of simplicity, the observation space is limited to a single image frame rather than multiple frame stacking.
Originally, the four tasks in Adroit have different action dimensions since certain tasks constrain the free movement of the robot hand. To learn a single multi-task policy, we set a unified action space as 6D movement of the root link and the 24-DoF joint control signal. 
Contrary to \cite{lfs}, which evaluates the policy at every epoch and selects the top three performances, we evaluate and report solely on the final policy's performance to avoid cherry-picking. 
It is crucial to note that these adjustments make our results not directly comparable to \cite{lfs}. However, by employing the same visual policy backbone across all experiments, we ensure fair comparisons between our framework and the baselines.

\subsection{Experimental Results}

This subsection aims to substantiate two key points:
1) AdaDemo can \textit{progressively improve} the performance of the multi-task visual policy across multiple rounds;
2) Compared to the uniform approach of data collection across all tasks and initial states, AdaDemo can expand the demonstration datasets in a more \textit{data-efficient} manner.

We implemented AdaDemo on the RLBench and Adroit benchmarks, with results presented in Fig. \ref{fig:brief_results}, Table \ref{tab:rlbench_result} and \ref{tab:adroit_result}. 
In these results, "Uniform" denotes the baseline method for expanding demonstrations, where demonstrations are evenly distributed across all tasks and initial states. 
Despite appearing simplistic, this uniform collection strategy is actually widely used in multi-task policy learning \cite{rvt, peract, james2022q}. 
For each round, we trained visual policies of the same architectures but utilized datasets collected separately by AdaDemo and the baseline, with the baseline dataset being comparable or slightly larger in size. The same demonstration collector was employed across both methods to ensure a fair comparison.

Fig. \ref{fig:brief_results} shows a clear trend: the average performance of the multi-task visual policy can be progressively improved by expanding the demonstration datasets.
Additionally, AdaDemo outperforms the baseline in two respects:
\begin{itemize}
    \item \textbf{Data Efficiency:} 
    AdaDemo demonstrates remarkable data efficiency, particularly at junctures where performance tends to plateau. It surpasses the baseline by requiring only 1/2 data for RLBench and 1/3 for Adroit. This efficiency could translate into significant cost reductions, especially in scenarios of large-scale data collection.
    \item \textbf{Performance Upper Bound:} 
    While the baseline method quickly hits performance plateaus, AdaDemo consistently improves the multi-task performance, broadening the performance gap between the two approaches with each round. This trend indicates a greater potential in performance upper bound as the demonstration dataset is scaling up.
\end{itemize}

Tables \ref{tab:rlbench_result} and \ref{tab:adroit_result} offer more detailed insights into performance differences. The RLBench results indicate that AdaDemo consistently outperforms the baseline in terms of average performance across tasks, although it does not lead in every single task. 
We hypothesize that the performance drop in certain RLBench tasks might be related to the issue of conflicting gradients in multi-task learning, as studied in \cite{yu2020gradient, shi2023recon, liu2021conflict}. AdaDemo's strategy of accumulating more demonstrations for harder tasks means that when gradients from these tasks do not align well with those from other tasks, following their gradient direction might adversely affect the performance on other tasks. Intuitively, focusing more on some tasks naturally detracts attention from others, and hence might hurt performance on other tasks.
This effect appears more pronounced in the RLBench experiments, where the gradients may be noisier due to a relatively small average number of demonstrations per task ($\sim200$), particularly impacting the easier tasks. Conversely, in the Adroit benchmark, which comprises a larger average number of demonstrations per task ($\sim3000$) and only four tasks, the issue of gradient conflict is potentially less severe. Here, AdaDemo outperforms the baseline across all tasks, with the Hammer task being an exception (given its relative ease, all learned policies perform nearly optimally).

We noted that the contribution to the overall performance improvement is concentrated in a few tasks, such as "Relocate" and "Door" in Adroit, as well as "Put in Cupboard" and "Stack Cups" in RLBench. These tasks, which are generally hard, significantly benefit from our adaptive demonstration expansion strategy. 
By allocating more demonstrations to these tasks, rather than to easier ones, AdaDemo enhances both \textit{data efficiency} and \textit{overall performance}.
However, it is important to note that the most difficult tasks (e.g., "Place Cups" and "Insert Peg") do not exhibit notable improvement with our adaptive demonstration expansion strategy. This is mainly due to these tasks being somewhat beyond the model's capabilities under the current setup. 
In such cases, exploring improvements to the model architecture or control method could be a worthwhile direction.

\subsection{Ablation Study}
\label{sec:ablation}

\begin{table}[t]
\vspace{0.2 cm}

\centering
\resizebox{\columnwidth}{!}{
\setlength\tabcolsep{2pt} %
\begin{tabular}{ccccc}
\toprule
\rowcolor[HTML]{D7DAFF}
                 & \textcolor{orange}{$\bullet$} Uniform & A & A+B & \begin{tabular}[c]{@{}c@{}}\textcolor{blue}{$\bullet$} AdaDemo\\ (A+B+C)\end{tabular} \\
\midrule
\rowcolor[HTML]{EFEFEF}
Put in Drawer    & 94.7{\footnotesize ±2.3}           & 80.0{\footnotesize ±10.6}          & \textbf{97.3}{\footnotesize ±4.6}  & 82.7{\footnotesize ±12.2}                                                   \\
Drag Stick       & 58.7{\footnotesize ±40.3}          & 84.0{\footnotesize ±20.8}          & 88.0{\footnotesize ±4.0}           & \textbf{92.0}{\footnotesize ±4.0}                                           \\
\rowcolor[HTML]{EFEFEF}
Turn Tap         & 93.3{\footnotesize ±2.3}           & 88.0{\footnotesize ±6.9}           & \textbf{94.7}{\footnotesize ±2.3}           & \textbf{96.0}{\footnotesize ±4.0}                                           \\
Slide Block      & 13.3{\footnotesize ±8.3}           & \textbf{21.3}{\footnotesize ±26.6} & 17.3{\footnotesize ±4.6}           & 17.3{\footnotesize ±6.1}                                                    \\
\rowcolor[HTML]{EFEFEF}
Open Drawer      & 74.7{\footnotesize ±4.6}           & 73.3{\footnotesize ±11.5}          & \textbf{81.3}{\footnotesize ±4.6}  & \textbf{80.0}{\footnotesize ±4.0}                                                    \\
Put in Cupboard  & 60.0{\footnotesize ±13.9}          & 64.0{\footnotesize ±10.6}          & 70.7{\footnotesize ±4.6}           & \textbf{77.3}{\footnotesize ±12.9}                                          \\
\rowcolor[HTML]{EFEFEF}
Sort Shape       & 37.3{\footnotesize ±8.3}           & \textbf{49.3}{\footnotesize ±2.3}  & 38.7{\footnotesize ±8.3}           & 41.3{\footnotesize ±10.1}                                                   \\
Put in Safe      & \textbf{93.3}{\footnotesize ±4.6}           & 90.7{\footnotesize ±6.1}           & \textbf{92.0}{\footnotesize ±8.0}           & \textbf{93.3}{\footnotesize ±2.3}                                           \\
\rowcolor[HTML]{EFEFEF}
Push Buttons     & \textbf{98.7}{\footnotesize ±2.3}  & 94.7{\footnotesize ±2.3}           & \textbf{97.3}{\footnotesize ±2.3}           & \textbf{96.0}{\footnotesize ±4.0}                                                    \\
Close Jar        & 48.0{\footnotesize ±13.9}          & 45.3{\footnotesize ±9.2}           & \textbf{58.7}{\footnotesize ±8.3}  & 48.0{\footnotesize ±6.9}                                                    \\
\rowcolor[HTML]{EFEFEF}
Stack Blocks     & 37.3{\footnotesize ±8.3}           & 42.7{\footnotesize ±4.6}           & \textbf{48.0}{\footnotesize ±12.0} & \textbf{46.7}{\footnotesize ±2.3}                                                    \\
Place Cups       & 4.0{\footnotesize ±4.0}   & 2.7{\footnotesize ±2.3}            & 0.0{\footnotesize ±0.0}            & 1.3{\footnotesize ±2.3}                                                     \\
\rowcolor[HTML]{EFEFEF}
Place Wine       & \textbf{92.0}{\footnotesize ±4.0}  & 89.3{\footnotesize ±9.2}           & 86.7{\footnotesize ±8.3}           & \textbf{90.7}{\footnotesize ±6.1}                                                    \\
Screw Bulb       & 62.7{\footnotesize ±8.3}           & 64.0{\footnotesize ±10.6}          & 65.3{\footnotesize ±8.3}           & \textbf{78.7}{\footnotesize ±4.6}                                           \\
\rowcolor[HTML]{EFEFEF}
Sweep to Dustpan & \textbf{57.3}{\footnotesize ±4.6}  & 42.7{\footnotesize ±12.9}          & 44.0{\footnotesize ±0.0}           & 45.3{\footnotesize ±2.3}                                                    \\
Insert Peg       & 16.0{\footnotesize ±4.0}           & \textbf{21.3}{\footnotesize ±2.3}           & \textbf{22.7}{\footnotesize ±4.6}  & 12.0{\footnotesize ±0.0}                                                    \\
\rowcolor[HTML]{EFEFEF}
Meat off Grill   & \textbf{100.0}{\footnotesize ±0.0} & \textbf{98.7}{\footnotesize ±2.3}           & 96.0{\footnotesize ±0.0}           & 94.7{\footnotesize ±4.6}                                                    \\
Stack Cups       & 45.3{\footnotesize ±20.5}          & 45.3{\footnotesize ±18.9}          & 48.0{\footnotesize ±6.9}           & \textbf{60.0}{\footnotesize ±6.9}                                           \\
\midrule
\rowcolor[HTML]{EFEFEF}
Average SR       & 60.4               & 61.0               & 63.7               & \textbf{64.1}                                               \\
Average \# Demo  & 200                & 191.2              & 191.2              & 191.2                                                       \\
\bottomrule
\end{tabular}
}

\caption{\textbf{Ablation Study:} A, B, C denote the 3 core principles in AdaDemo, see Sec. \ref{sec:ablation} for a detailed explanation. }
\label{tab:ablation_rlbench}

\end{table}

We conduct an ablation study to examine the importance of the three guiding principles (Sections \ref{sec:demo_on_failed}, \ref{sec:demo_on_hard}, and \ref{sec:sample_startegy}) underlying AdaDemo. 
These principles, denoted as A, B, and C, are delineated as follows:

\begin{itemize}
    \item A: Collecting demonstrations on unsolved tasks.
    \item B: Collecting demonstrations on failed initial states.
    \item C: Sampling data uniformly with respect to the demonstration trajectories.
\end{itemize}

To evaluate the influence of each principle, we start with the variant that uses none of these principles, and add one principle at one time, to create four variants. 
This ablation study allows us to dissect the specific role and importance of each component towards improving a multi-task visual policy, utilizing a comparable amount of data.
The ablated variants are:

\begin{itemize}
    \item \textbf{None of ABC:} This variant is the same as the "Uniform" baseline mentioned above, where demonstrations are uniformly collected across all tasks and initial states, without the targeted strategies of AdaDemo.
    \item \textbf{A}: Here, we collect demonstrations across tasks with the same number as in our full framework, but unlike our proposed approach, these demonstrations are sourced from all initial states, not just the failed ones. 
    \item \textbf{A + B:} This variant combines the collection of more demonstrations on unsolved tasks (A) with the focus on failed initial states (B). However, unlike our full framework, it employs uniform sampling with respect to tasks, not demonstration trajectories. 
    \item \textbf{A + B + C}: Our complete framework incorporates all three principles, providing a reference to compare against the ablated variants.
\end{itemize}

All the variants are applied on RLBench, and the results are shown in Table \ref{tab:ablation_rlbench}. Analyzing the experimental results, we observe that each principle contributes to the overall performance gain, albeit to varying degrees. Notably, principle B, which emphasizes collecting demonstrations on failed initial states, appears to have a more substantial impact than principles A (focusing on unsolved tasks) and C (uniform sampling w.r.t. demonstration trajectories). This finding highlights the critical role of targeted demonstration collection in areas where the policy currently fails, underscoring the importance of addressing specific weaknesses in the policy for effective learning.

\section{Limitations}

We would like to discuss a few limitations of our work. 
Firstly, AdaDemo requires policy evaluation to determine the scenarios for collecting new demonstrations, and this cost is not precisely accounted for in comparison to baselines. Nevertheless, it is important to note that the primary cost arises from the demonstration collection itself, which potentially involves human effort. Policy evaluation, by contrast, is a more autonomous procedure, and its costs are likely to be much lower than those incurred by demonstration collection.
Secondly, AdaDemo assumes the availability of a success metric for tasks, a condition readily met in simulations but potentially hard in the real world. However, obtaining success signals in the real world is still feasible, and it has been achieved by methods like visual detection/tracking \cite{qt_opt} or employing large vision-language models \cite{du2023vision}.

\section{Discussions and Conclusions}

This study introduces AdaDemo, a novel framework designed to actively and continuously expand the demonstration dataset for training generalist robotic agents.
AdaDemo can be categorized as a form of active learning tailored for robotics. Active learning methods have seen broad application in robotics, spanning active sensing \cite{thuruthel2019soft, sornkarn2016can}, active coverage \cite{paull2012sensor}, and active SLAM \cite{cadena2016past}. Our research aligns with the spirit of active learning, focusing on efficiently collecting additional demonstrations by a demonstration collector that only handles initial task states.
A standout feature of AdaDemo is its exceptional data efficiency, especially evident at points where the performance of the policy reaches a plateau. This efficiency holds the potential for considerable cost reductions, especially in scenarios of large-scale demonstration collection.

\bibliographystyle{ieeetr}
\bibliography{refs}

\newpage

\section*{APPENDIX}

\section{Implementation Details}

\subsection{RVT on RLBench}

In our RLBench experiments, we employed RVT \cite{rvt} as the foundational visual policy. While adhering to most aspects of the original RVT implementation, we introduced some modifications to address a \href{https://github.com/peract/peract/issues/6#issuecomment-1355555980}{legacy issue} that resulted in unnecessarily large training datasets. Optimizing storage was crucial, especially considering that our framework expands the demonstration dataset. We detail our three key modifications below:

\begin{enumerate}
    \item In the original RVT implementation (and its predecessor \cite{peract, james2022q}), each keyframe was redundantly added to the dataset $k$ times, with $k$ being the keyframe's index. This redundancy, acknowledged as a leftover artifact by the authors of \cite{peract}, was eliminated in our approach. We revised the process to include each keyframe only once in the dataset, effectively reducing the size of the resulting dataset.

    \item The removal of duplicate keyframes initially led to a performance drop in RVT. We found a common error pattern of the robot getting stuck at certain keyframes. We hypothesized that this problem was due to the training dataset's structure of ${(s,a), ...}$, where $s$ represents the current state, and $a$ is the end-effector pose for \textbf{the next keyframe}. If the current frame is too close to the next keyframe, the policy tends to yield minimal end-effector movement, causing the robot to stay still. To address this issue, we developed a new approach for assigning target end-effector poses. For frames between keyframes $k$ and $k+1$, the target pose is determined based on proximity: if closer to $k+1$, the target is set from keyframe $k+2$, otherwise from $k+1$. By our empirical evaluation, this adjustment significantly mitigated the "stuck" behavior and enhanced overall performance.

    \item Additionally, to further diminish the occurrence of the robot getting stuck, we modified the timestep encoding in the dataset. The original RVT augmented each frame with a timestep dimension, linearly decaying from $1$ to $-1$ based on a predefined maximum episode length. We randomized timestep encoding during both training and evaluation, adding noises to help the agent break out of potential loops.
\end{enumerate}

Please note that the effectiveness of the aforementioned modifications has only been validated empirically. However, we do not claim these modifications as the contributions of our work.

As a result of these three modifications, our adapted version of RVT is more storage-efficient, with only a minor impact on performance. Crucially, the consistent application of this modified algorithm across all RLBench experiments ensures a fair comparison between our framework and the baselines.

When collecting demonstrations from the failed initial states of each task, we set an upper limit of 100, 200, and 200 new demonstrations for each round, respectively. The rationale behind this is to prevent the potential indefinite duration required to attain a targeted number of successful episodes for some tasks with extremely low success rates.

\subsection{LfS on Adroit}

In our experiments on Adroit tasks, we followed \cite{lfs}, employing a simple convolutional neural network with random shifting augmentation as our visual policy backbone. While \cite{lfs} limited their experiments to only two Adroit tasks (pen and relocate), we expanded our scope to include all four Adroit tasks: pen, relocate, hammer, and door. While retaining most of the implementation aspects from \cite{lfs}, we made a few adjustments, detailed as follows:

\begin{itemize}
    \item In the original Adroit setup, each task had varying action dimensions, primarily because certain tasks restricted the Adroit hand's free movement in 3D space. To facilitate multi-task learning, we unified the action space to include 6D movement of the root link and a 24-DoF joint control signal. Note that we did not modify the tasks themselves, and the extra action dimensions for certain tasks will be ignored in the evaluation.
    \item For simplicity, our observation space utilizes only a single frame of images rather than stacking multiple frames.
    \item The evaluation approach in \cite{lfs} involves averaging results from the top-3 performing checkpoints out of 50. To eliminate potential bias from such selective reporting, we assess and report performance based solely on the final checkpoints.
\end{itemize}

It is important to note that due to these modifications, our results are not directly comparable with those in \cite{lfs}. Nonetheless, by using the same visual policy backbone for both our framework and the baselines, we still maintain fairness in the comparisons presented in our paper.

When collecting demonstrations from the failed initial states of each task, we set an upper limit of 40, 120, 400, 1400, and 6000 new demonstrations for each round, respectively. This cap is determined as twice the average number of demonstrations targeted for collection in each corresponding round. The rationale behind this is to prevent the potential indefinite duration required to attain a targeted number of successful episodes for some tasks with extremely low success rates.

\newpage

\onecolumn

\section{Num of Trajectories for Each Task}

Our framework iteratively expands the demonstration dataset, and these demonstrations are not uniformly distributed across each task. In this section, we present the number of trajectories for each task in the demonstration dataset, as accumulated in each round of our framework. The numbers are shown in Table \ref{tab:rlbench_num_demo} and \ref{tab:adroit_num_demo}.

\begin{table*}[ht]
\vspace{1cm}
\centering
\begin{tabular}{c|c|cc|cc|cc}
\toprule
                 & Initial & \multicolumn{2}{c|}{Round 1} & \multicolumn{2}{c|}{Round 2} & \multicolumn{2}{c}{Round 3} \\
                 &         & \textcolor{blue}{$\bullet$} AdaDemo         & \textcolor{orange}{$\bullet$} Uniform        & \textcolor{blue}{$\bullet$} AdaDemo         & \textcolor{orange}{$\bullet$} Uniform        & \textcolor{blue}{$\bullet$} AdaDemo         & \textcolor{orange}{$\bullet$} Uniform        \\
\midrule
Put in Drawer    & 20      & 21          & 50             & 31          & 100            & 106         & 200           \\
Drag Stick       & 20      & 21          & 50             & 31          & 100            & 49          & 200           \\
Turn Tap         & 20      & 20          & 50             & 21          & 100            & 41          & 200           \\
Slide Block      & 20      & 54          & 50             & 144         & 100            & 344         & 200           \\
Open Drawer      & 20      & 21          & 50             & 24          & 100            & 39          & 200           \\
Put in Cupboard  & 20      & 73          & 50             & 98          & 100            & 177         & 200           \\
Sort Shape       & 20      & 36          & 50             & 93          & 100            & 246         & 200           \\
Put in Safe      & 20      & 22          & 50             & 29          & 100            & 55          & 200           \\
Push Buttons     & 20      & 26          & 50             & 29          & 100            & 40          & 200           \\
Close Jar        & 20      & 38          & 50             & 98          & 100            & 231         & 200           \\
Stack Blocks     & 20      & 86          & 50             & 186         & 100            & 386         & 200           \\
Place Cups       & 20      & 120         & 50             & 220         & 100            & 420         & 200           \\
Place Wine       & 20      & 22          & 50             & 23          & 100            & 69          & 200           \\
Screw Bulb       & 20      & 43          & 50             & 77          & 100            & 190         & 200           \\
Sweep to Dustpan & 20      & 26          & 50             & 76          & 100            & 245         & 200           \\
Insert Peg       & 20      & 120         & 50             & 220         & 100            & 420         & 200           \\
Meat off Grill   & 20      & 21          & 50             & 22          & 100            & 24          & 200           \\
Stack Cups       & 20      & 120         & 50             & 220         & 100            & 359         & 200           \\
\midrule
Average \# Demo  & 20      & 49.4        & 50             & 91.0        & 100            & 191.2       & 200           \\
\bottomrule
\end{tabular}
\caption{RLBench: Number of trajectories for each task in demonstration datasets.}
\label{tab:rlbench_num_demo}
\end{table*}

\begin{table*}[ht]
\centering
\resizebox{\columnwidth}{!}{
\setlength\tabcolsep{2pt}
\begin{tabular}{c|c|cc|cc|cc|cc|cc}
\toprule
                & Initial  & \multicolumn{2}{c|}{Round 1} & \multicolumn{2}{c|}{Round 2} & \multicolumn{2}{c|}{Round 3} & \multicolumn{2}{c|}{Round 4} & \multicolumn{2}{c}{Round 5}  \\
                &          & {\small \textcolor{blue}{$\bullet$} AdaDemo}         & {\small \textcolor{orange}{$\bullet$} Uniform}      & {\small \textcolor{blue}{$\bullet$} AdaDemo}         & {\small \textcolor{orange}{$\bullet$} Uniform}      & {\small \textcolor{blue}{$\bullet$} AdaDemo}         & {\small \textcolor{orange}{$\bullet$} Uniform}      & {\small \textcolor{blue}{$\bullet$} AdaDemo}              & {\small \textcolor{orange}{$\bullet$} Uniform} & {\small \textcolor{blue}{$\bullet$} AdaDemo}              & {\small \textcolor{orange}{$\bullet$} Uniform }  \\
\midrule
Relocate        & 20      & 60          & 40            & 180         & 100           & 580         & 300           & 1980        & 1000 & 7980        & 3000           \\
Door            & 20      & 35          & 40            & 79          & 100           & 206         & 300           & 922         & 1000 & 1903        & 3000           \\
Pen             & 20      & 36          & 40            & 116         & 100           & 353         & 300           & 1060        & 1000   & 2068        & 3000         \\
Hammer          & 20      & 20          & 40            & 21          & 100           & 21          & 300           & 26          & 1000   & 26        & 3000         \\
\midrule
Average \# Demo & 20      & 37.8        & 40            & 99.0        & 100           & 290.0       & 300           & 997.0       & 1000  & 2994.3        & 3000     \\
\bottomrule
\end{tabular}
}
\caption{Adroit: Number of trajectories for each task in demonstration datasets.}
\label{tab:adroit_num_demo}
\end{table*}

\end{document}